\documentclass[conference]{IEEEtran}
\IEEEoverridecommandlockouts

\usepackage{url}

\usepackage{breakurl}
\usepackage[breaklinks]{hyperref}

\usepackage[papersize={8.5in,11in},
            top=1in,
            left=0.75in,
            right=0.75in,
            bottom=0.75in]{geometry}

\makeatletter
\def\endthebibliography{%
	\def\@noitemerr{\@latex@warning{Empty `thebibliography' environment}}%
	\endlist
}
\makeatother

\usepackage{stfloats}
\usepackage{tabularx,booktabs}
\usepackage{cite}
\usepackage{authblk}
\usepackage{amsmath,amssymb,amsfonts}
\usepackage{algorithmic}
\usepackage{graphicx}
\usepackage{textcomp}
\usepackage{xcolor}
\def\BibTeX{{\rm B\kern-.05em{\sc i\kern-.025em b}\kern-.08em
    T\kern-.1667em\lower.7ex\hbox{E}\kern-.125emX}}
\begin{document}

\title{Vectorized Representation Dreamer (VRD): Dreaming-Assisted Multi-Agent Motion Forecasting  \\
}

\author[1, 2]{Hunter Schofield}
\author[1]{Hamidreza Mirkhani}
\author[1]{Mohammed Elmahgiubi}
\author[1]{Kasra Rezaee}
\author[2]{Jinjun Shan}

\affil[1]{Noah's Ark Lab, Huawei Technologies Canada}
\affil[2]{York University}
\affil[ ]{\textit {\{hunter.schofield,hamidreza.mirkhani,mohammed.elmahgiubi,kasra.rezaee\}@huawei.com}}
\affil[ ]{\textit {\{hunterls,jjshan\}@yorku.ca}}


\maketitle

\begin{abstract}
For an autonomous vehicle to plan a path in its environment, it must be able to accurately forecast the trajectory of all dynamic objects in its proximity. While many traditional methods encode observations in the scene to solve this problem, there are few approaches that consider the effect of the ego vehicle's behavior on the future state of the world. In this paper, we introduce VRD, a vectorized world model-inspired approach to the multi-agent motion forecasting problem. Our method combines a traditional open-loop training regime with a novel dreamed closed-loop training pipeline that leverages a kinematic reconstruction task to imagine the trajectory of all agents, conditioned on the action of the ego vehicle.
Quantitative and qualitative experiments are conducted on the Argoverse 2 multi-world forecasting evaluation dataset and the intersection drone (inD) dataset to demonstrate the performance of our proposed model. Our model achieves state-of-the-art performance on the single prediction miss rate metric on the Argoverse 2 dataset and performs on par with the leading models for the single prediction displacement metrics. 
\end{abstract}

\section{Introduction}
To navigate to a destination, an autonomous vehicle must be capable of accurately predicting the behavior of other dynamic objects which also share the road. Then, the autonomous vehicle must plan a trajectory for itself that does not violate traffic rules or compromise safety. Many approaches to this problem aim to generate trajectories that are close to a ground truth expert trajectory and do not intersect the trajectories of other surrounding dynamic objects \cite{Lee_desire}, \cite{Fang_tpnet}. These approaches are well suited to finding trajectories that are similar to experts, but this does not necessarily translate to an improved planning performance when the trajectory is unrolled in a closed loop. One way to solve this problem is to combine techniques from imitation learning (IL) and reinforcement learning (RL), relying on IL for scenarios that are in the training data distribution and RL for when a vehicles' observations go out of distribution \cite{Lu_not_enough}.  However, this requires designing a cost function that accurately captures our objectives as human drivers, which is difficult to do \cite{Knox_misdesign}. A more direct approach to solving this problem is to do training directly in the closed loop, which is done in state-of-the-art methods such as model-based adversarial imitation learning (MGAIL) \cite{Baram_MGAIL}, \cite{Bronstein_HMGAIL}. In MGAIL, vehicle dynamics are simulated over some horizon, and a discriminator is used to distinguish expert and novice states. At the end of the horizon, backpropagation through time (BPTT) is used to update the policy. Following this inspiration, in this paper we propose Vectorized Representation Dreamer (VRD), a  model-based approach that solves the multi-agent motion forecasting problem by learning a transition function that operates on the latent world representation and ego action, which enables a realistic dream of the future that is conditioned on the behavior of the ego vehicle. Fig. \ref{fig:vrd_illustration} illustrates the trajectory forecasting in the dreamed environment by VRD. The contributions of this paper are:
\begin{figure}[!t]
    \centering
    \includegraphics[scale=0.25]{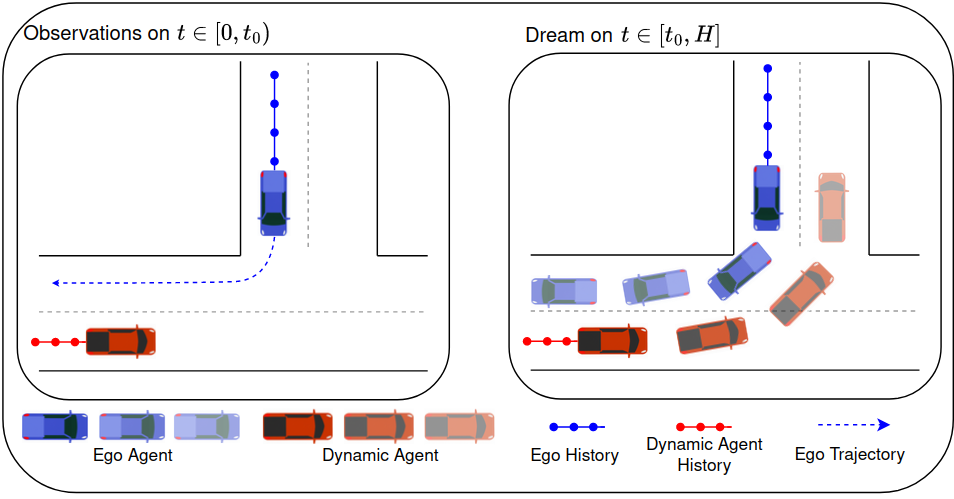}
    \caption{Illustration of our motion forecasting model. Based on the historical trajectories of all objects, the ego agent plans an initial trajectory. Using this trajectory, the ego dreams the future of all dynamic objects in the environment. Since the historical observation of the red car indicates slow movement, the model can infer that the red car is likely turning instead of going straight.}
    \label{fig:vrd_illustration}
    \vspace{-3mm}
\end{figure}

\vspace{-1mm}
\begin{itemize}
    \item The development of a novel motion forecasting framework, VRD, that learns a world model on vectorized representations of the environment.
    \item The introduction of the kinematic state reconstruction task that helps the latent space better capture aspects of the environment important to motion forecasting of dynamic objects.
    \item A new training method for motion forecasting that combines traditional open-loop training with an imagined closed-loop training regime carried out within the dreamed future of the model. 
\end{itemize}

Based on the imagined future scenario representation, we predict a trajectory for all agents in the environment that performs on par or better than other state-of-the-art methods when evaluated on the Argoverse 2 motion forecasting benchmark. The paper is organized as follows. Section \ref{s2} introduces the related work. Section \ref{s3} outlines the VRD pipeline for generating agent trajectories and the open and closed-loop training regimes. Section \ref{s4} provides qualitative and quantitative results regarding the multi-agent trajectory generation. Section \ref{s5} overviews the conclusions of this work.

\section{Related Work \label{s2}}
\subsection{Motion Forecasting}
The popularity of motion forecasting has been increasing in recent years following the introduction of large-scale motion forecasting datasets such as nuScenes \cite{Caesar_Nuscenes}, the Waymo open motion dataset \cite{WOMD}, INTERACTION \cite{Zhan_Interaction}, and Argoverse 2 motion forecasting \cite{Argoverse2}. Recent works such as HiVT \cite{Zhou_HIVT} explore a hierarchical approach to vectorized scene encodings to model large numbers of dynamic objects efficiently. HeteroGCN \cite{Gao_HeteroGCN} uses a graph convolutional network to capture relationships between multiple nodes in a dynamic graph that models each scenario. Approaches like Forecast-MAE \cite{Cheng_ForecastMAE} use a masked autoencoder to randomly remove either the history or future trajectory of the object which a decoder then tries to reconstruct. The current state of the art approach, QCNet \cite{Zhou_QCNet}, uses Fourier embeddings of object trajectories in polar coordinates to learn high-frequency signals and then uses self-attention modules to encode map information. These embeddings provide an encoding of the environment that is invariant under transformations of a global frame of reference.

\subsection{Imitation Learning}
The goal of imitation learning is to train a policy that can predict an expert's behaviour using data that couples an expert's observation and action. While a learned policy may succeed in accurately predicting the expert's action from a specific observation, since future observations are conditioned on the previous actions, any error made will propagate forward causing future observations to become out of distribution from the training dataset \cite{Ross_Dagger}. This is known as the covariate shift problem that open-loop IL approaches such as behaviour cloning will suffer from \cite{Pomerleau_ALVINN}, \cite{Salzmann_trajectron}, \cite{Ngiam_scene}. To alleviate this problem state-of-the-art methods extend IL training to closed-loop simulations using generative adversarial networks (GANs) \cite{Bronstein_HMGAIL}, \cite{Keufler_GAN}. Another state-of-the-art approach, model-based imitation learning (MILE) \cite{Hu_Mile} does behaviour cloning in the closed loop, and tries to learn a policy that produces similar actions to an expert based on an imagined future. Another approach is to use imitation learning to break the problem into separate learnable tasks such as target prediction, motion estimation, and trajectory scoring as is done in Target-driveN Trajectory (TNT) \cite{Zhao_TNT}.

\subsection{World Representation}
High-definition (HD) maps can be used to imbue an agent with topographical information, allowing autonomous vehicles to choose sensible actions based on the environment they operate in \cite{Yang_HDnet}. To learn useful information from HD maps to learn better driving policies, approaches like ChauffeurNet \cite{Bansal_ChauffeurNet} rasterize an HD map to create masks that capture individual map features such as lanes, dynamic objects or the desired route. 2D Convolutional networks are then used to extract the important features from these masks. However, rasterization is a lossy process and results in images that contain regions with unhelpful information. Instead, approaches like LaneGCN \cite{Liang_LaneGCN} and VectorNet \cite{Gao_Vectornet} encode relationships between map features into graphs and then use graph neural networks (GNN) to extract a latent feature vector from the maps. 

\subsection{World Models}
World models have had remarkable success in learning latent dynamics models that can infer the future latent representations of the world \cite{Ha_WorldModels}. Many approaches to learned world models try to learn a reward predictor to train a policy directly in the imagined environment \cite{Hafner_Learn2Control}, \cite{Hafner_dreamerv2} or to train a policy that can select the best action based on a tree of all possible future states within some temporal horizon \cite{Schrittwieser_muzero}. More recently, world models have been used to learn realistic driving simulators \cite{Zhang_Trafficbots} and conduct imitation learning \cite{Hu_Mile}.

\section{Methodology \label{s3}}
\begin{figure*}[ht!]
    \centering
    \includegraphics[scale=0.75]{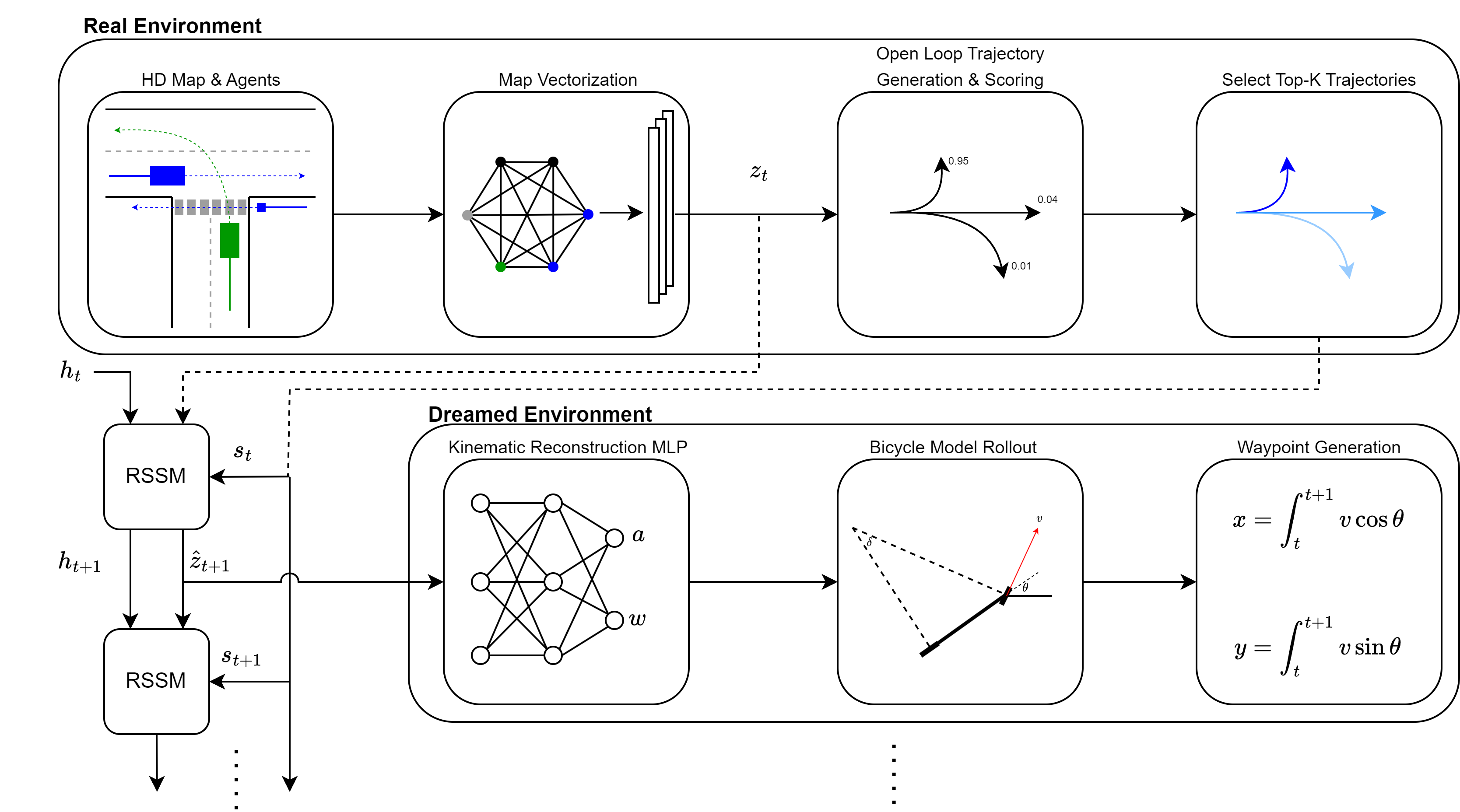}
    \caption{VRD pipeline overview. All map features and objects are processed into a vectorized latent space, $z_t$. Then, a dreamed rollout is produced by passing the trajectory along with the previous latent representation to the RSSM. The transition predictor estimates the next latent representation of the world which is decoded to obtain the kinematic states of all agents. This process is iterated to re-plan a new ego trajectory, closing the dreamed loop.}
    \label{fig:vrd_pipeline}
    \vspace{-5mm}
\end{figure*}

We aim to learn a robust driving policy from a vectorized latent representation of the dynamic objects in an environment. The proposed imitation learning framework leverages a recurrent state-space model (RSSM) which predicts the next latent state of the world from the prior latent state and ego vehicle action. We also propose a kinematic state reconstruction header network learn vehicle kinematics from the latent representation. This allows the framework to unroll vehicle kinematics over some horizon, $H$. Evaluating the model in this imagined horizon gives closed-loop feedback, and leveraging suitable loss functions on these imaginations can improve the model's closed-loop performance via BPTT. Details of these components are provided in the following sections.

\subsection{World Model}
The design of our world model is inspired by the RSSM proposed by Hafner \emph{et al}. \cite{Hafner_dreamerv2}. In our design however, since we are trying to learn a transition function for the vectorized latent space representation, VectorNet \cite{Gao_Vectornet} is used as the representation model. Furthermore, since we are interested in training an accurate representation of the future, our transition model is deterministic instead of the variational structure proposed in \cite{Hafner_dreamerv2}.  Thus, our RSSM can be summarized by the three key networks, the recurrent model, $f$, the representation model, $g$, and the transition predictor, $p$:

\vspace{-3mm}
\begin{equation}
    RSSM = \begin{cases}
        h_t = f_{\phi}(h_{t-1}, z_{t-1}, a_{t-1}) \\
        z_t = g_{\phi}(x_t) \\
        \hat{z}_t = p_{\phi}(h_t)
    \end{cases}
\end{equation}

\noindent
where $h$ is the hidden state of the recurrent model, $z$ is the vectorized latent world representation, $\hat{z}$ is the predicted vectorized latent world representation from the priors, $a$ is the ego action, and the RSSM is parameterized by $\phi$. The recurrent model is implemented as a gated recurrent unit (GRU) with a single cell, and the transition predictor is implemented as a feed-forward network with a skip connection between the first and final layers. A modified cosine similarity loss criterion is used to train the world model prediction. At this point, the gradients on $\hat{z}_t$ are detached so that the transition predictor moves towards the representation model and not vice versa.

\vspace{-3mm}
\begin{equation}
    \mathcal{L}_{RSSM} = 1 - \frac{z_t \cdot \hat{z}_t}{ \max \left( |z_t|, |\hat{z}_t| \right)}
\end{equation}

\subsection{Kinematic State Reconstruction}
At each time step, the latent representation produced by VectorNet is decoded by a kinematic header network to predict the acceleration, $a$, and rate of turn, $w$, of each vehicle in the scene. This header network is implemented as a 2 layer MLP. The position and heading of each vehicle can then be determined by solving the bicycle model kinematic equations. The velocity is determined by integrating the predicted acceleration. Similarly, the vehicle yaw can be determined by integrating the rate of turn and then the $x$ and $y$ coordinates can be determined using standard kinematic equations.

\vspace{-1mm}
\subsection{Open Loop Training}
The imitation learning module begins by predicting a set of $N$ discrete targets that the ego vehicle is likely to reach, following the TNT target prediction structure.

\begin{equation}
    \mathcal{T} = \{ (x^n, y^n) + (\Delta x^n, \Delta y^n) \}_{n=1}^N
\end{equation}

A target distribution can be created using two multi-layer perceptrons (MLP), $u$ and $v$ operating on the latent representation, $z$, provided by the representation model.

\begin{equation}
    p(\tau^n | z) = \frac{e^{u(\tau^n | z)}}{\sum_{\tau'}u(\tau' | z)} \cdot \mathcal{N}(\Delta x^n | v_x^n(z)) \cdot \mathcal{N}(\Delta y^n | v_y^n(z))
\end{equation}

\noindent
At the next stage of open loop training, the top $M$ targets are sampled from the target prediction network. An MLP network then takes these targets along with the latent representation to produce a trajectory $\hat{S}_T = \left[ \hat{s}_0, ..., \hat{s}_T \right]$ for each target. Following the final stage of the TNT structure, a maximum entropy model is used to score each trajectory:

\begin{equation}
    \phi(s|z) = \frac{e^{g(s, z)}}{\sum_{m=1}^{M}e^{g(s_m, z)}}
\end{equation}

\noindent
here, $g(\cdot)$ is a MLP. The loss for this scoring module is a cross-entropy between the predicted scores and a ground truth score that is defined based on the maximum distance between all coordinates in the predicted trajectory and their temporally corresponding coordinate in the ground truth trajectory.

\subsection{Closed Loop Training in Imagination}
Once each trajectory is scored, the top-scoring trajectory is selected to be unrolled by the transition model. The closed loop training regime begins by passing the vectorized latent space, $z_t$, as well as the first timestep of the highest-scored trajectory, $s_t$ to the transition model to get the predicted vectorized latent space at the next time step, $\hat{z}_{t+1}$. The imagined latent space is then passed to the kinematic reconstruction header MLP to get the acceleration and steering rate of each dynamic object in the scene, including the ego vehicle. A bicycle model is then used to estimate the position of each object using the prior state information and reconstructed kinematics. At the same time, the imagined latent space and the original target prediction are passed again to the ego trajectory generation module to produce a new ego trajectory starting from the next time step. This new trajectory is fed back to the transition model with the newly predicted latent representation to close the dreamed loop. We iterate this process until the temporal horizon, $H$, is reached. Our model pipeline is illustrated in Fig. \ref{fig:vrd_pipeline}. To ensure that the dream learns a suitable forecast for all dynamic objects, a smooth L1 loss is introduced to force the dreamed trajectories to converge towards the ground truth trajectories over the imagined horizon:

\vspace{-1mm}
\begin{equation}
    \mathcal{L}_{S} = \sum_{t=0}^{H} \begin{cases}
        \frac{1}{2} ||\hat{s}_{t} - s_{t}||_2 & ||\hat{s}_{t} - s_{t}||_1 < 1 \\
        ||\hat{s}_{t} - s_{t}||_1 - \frac{1}{2} & \textrm{otherwise}
    \end{cases}
\end{equation}

\vspace{-1mm}
\noindent
where $\hat{s}_{t} = \begin{bmatrix} x & y & a & w \end{bmatrix}^T$ is the dreamed trajectory state at horizon step $t$, $s_{t}$ is the ground truth trajectory state at horizon step $t$, and $||\cdot||_1$ and $||\cdot||_2$ are the l1 and l2 norms, respectively.

\vspace{-1mm}
\section{Experimental Results \label{s4}}
We use the Argoverse 2 \cite{Argoverse2} motion forecasting dataset, and the intersection drone (inD) \cite{inDdataset} dataset to test our approach. These datasets contain a large variety of object classes, which provides a suitable benchmark for evaluating the ability of our transition model to capture latent representations for a wide range of vehicle dynamics. The Argoverse 2 dataset contains 250,000 scenarios with 10 object classes across 6 different cities and sampled at a rate of 10 Hz. Each scenario in the Argoverse 2 dataset lasts 11 seconds, with a 5-second observation window and a 6-second prediction horizon. The inD dataset uses a drone to record intersection scenarios containing 5 object classes across 4 locations. Scenarios on the inD dataset are longer than the Argoverse 2 dataset, so we also segment the inD scenarios to create samples of the same length as found in Argoverse 2. Unlike the Argoverse 2 dataset, the InD dataset solely contains intersection scenarios where frequent interactions occur between the dynamic objects. This provides a suitable benchmark for evaluating how well our transition model learns highly interactive behaviors, such as yielding to pedestrians, instead of just learning an accurate kinematic estimation. Since our transition model needs to be trained on samples with the same size observation window, we only use a 4-second observation window to make predictions. The metrics used to evaluate the models are the minimum average displacement error, $\mathrm{minADE_K}$, the minimum final displacement error, $\mathrm{minFDE_K}$, and the miss rate, $\mathrm{actorMR_K}$, where $K \in \{1, 6\}$ represents the number of trajectories that are predicted for each sample. For the displacement metrics, $\mathrm{minADE_K}$, is the minimum Euclidean norm across $K$ predictions between the predicted trajectory and the ground truth trajectory, when averaged over the prediction horizon for all objects in the environment. The $\mathrm{minFDE_K}$ metric is the minimum displacement error across $K$ predictions between the final predicted coordinate and final ground truth coordinate for all objects in the environment. The $\mathrm{actorMR_K}$ metric is the fraction of predictions with a minimum final displacement error of less than 2 meters across $K$ trajectories. Since our predictions are based on a deterministic recurrent state-space model, the dreamed future is also deterministic. Due to this, we only evaluate our model against the metrics for $K=1$. Table \ref{tab:argov2_results}. compares VRD trained on 70 epochs of the Argoverse 2 training split with the current state-of-the-art methods on the single predictions benchmarks. 

\begin{table}[!t]
    \centering
    \caption{Argoverse 2 Model Comparison}
    \begin{tabular}{c|c|c|c}
        \hline\hline
        Model & $\mathrm{minADE_1 \downarrow}$ & $\mathrm{minFDE_1 \downarrow}$ & $\mathrm{actorMR_1 \downarrow}$   \\
        \hline
        QML \cite{Su_QML} & 1.84 & 4.98 & 0.62  \\
        \hline
        HeteroGCN \cite{Gao_HeteroGCN} & 1.72 & 4.40 & 0.59 \\
        \hline
        Forecast-MAE \cite{Cheng_ForecastMAE} & 1.66 & 4.14 & 0.59 \\
        \hline
        QCNet \cite{Zhou_QCNet} & \textbf{1.56} & \textbf{3.96} & 0.55  \\
        \hline
        VRD (Ours) & 2.19 & 5.68 & \textbf{0.36} \\
        \hline\hline
        Forecast-MAE \cite{Cheng_ForecastMAE} & 1.76 & 4.39 & 0.59 \\
        \hline
        QCNet \cite{Zhou_QCNet} & 1.69 & 4.32 & 0.58  \\
        \hline
        VRD (Ours) & \textbf{1.68} & \textbf{4.17} & \textbf{0.37} \\
        \hline\hline
    \end{tabular}
    \vspace{1mm}
    \begin{flushleft}
    Table \ref{tab:argov2_results}. Comparison with state-of-the-art results on the Argoverse 2 multi-world motion forecasting test (top) and validation (bottom) datasets. The best results are \textbf{bolded}. Our world model based trajectory generation framework performs on par or better than some of the leading motion forecasting models.
    \end{flushleft}
    \label{tab:argov2_results}
    \vspace{-10mm}
\end{table}

The results from Table \ref{tab:argov2_results} demonstrate that our dreaming-assisted approach to motion forecasting is comparable to current state-of-the-art models on the displacement metrics, and outperforms the current state-of-the-art models on the single prediction miss rate metric. We chose these models to compare against since they each employ a different architecture to encode environmental features as highlighted in section \ref{s2}. Likewise, our architecture in VRD is also a novel approach as we train a transition function to predict the evolution of the scene features conditioned on the ego action. Since each dataset contains many different classes of ego objects, and each road user has a different driving style, the transition model can learn different driving behaviors by learning how each action changes the latent representation of the scene. Thus, given a short observation window, the transition model dreams a kinematic rollout following the observed driving behavior that accurately represents the future trajectory. Specifically, the percentage of predictions that have a final coordinate more than 2 meters from the ground truth is 36\% less than QCNet, demonstrating that VRD achieves state-of-the-art performance on the single prediction miss rate metric. 
VRD outperforms on the miss rate metric, however, it has a slightly higher $\mathrm{minADE_1}$ and $\mathrm{minFDE_1}$ compared to the other benchmark models. The main reason for this is due to a catastrophic distribution shift that can occur during the dreamed rollout. In these cases, the first few frames of the dream may be accurate, but over a sufficiently long horizon, the predicted latent representation begins to diverge. Thus, the reconstructed kinematics are effectively random, and the difference between the predicted trajectories and ground truth trajectories can be large. However, since the miss rate metric does not consider the scale of the displacement error, it is minimally impacted by this failure. An example of this type of failure is illustrated in the open loop prediction failure case of \cite{think2drive}.


\begin{figure*}[!ht]
    \centering
    \includegraphics[scale=0.105]{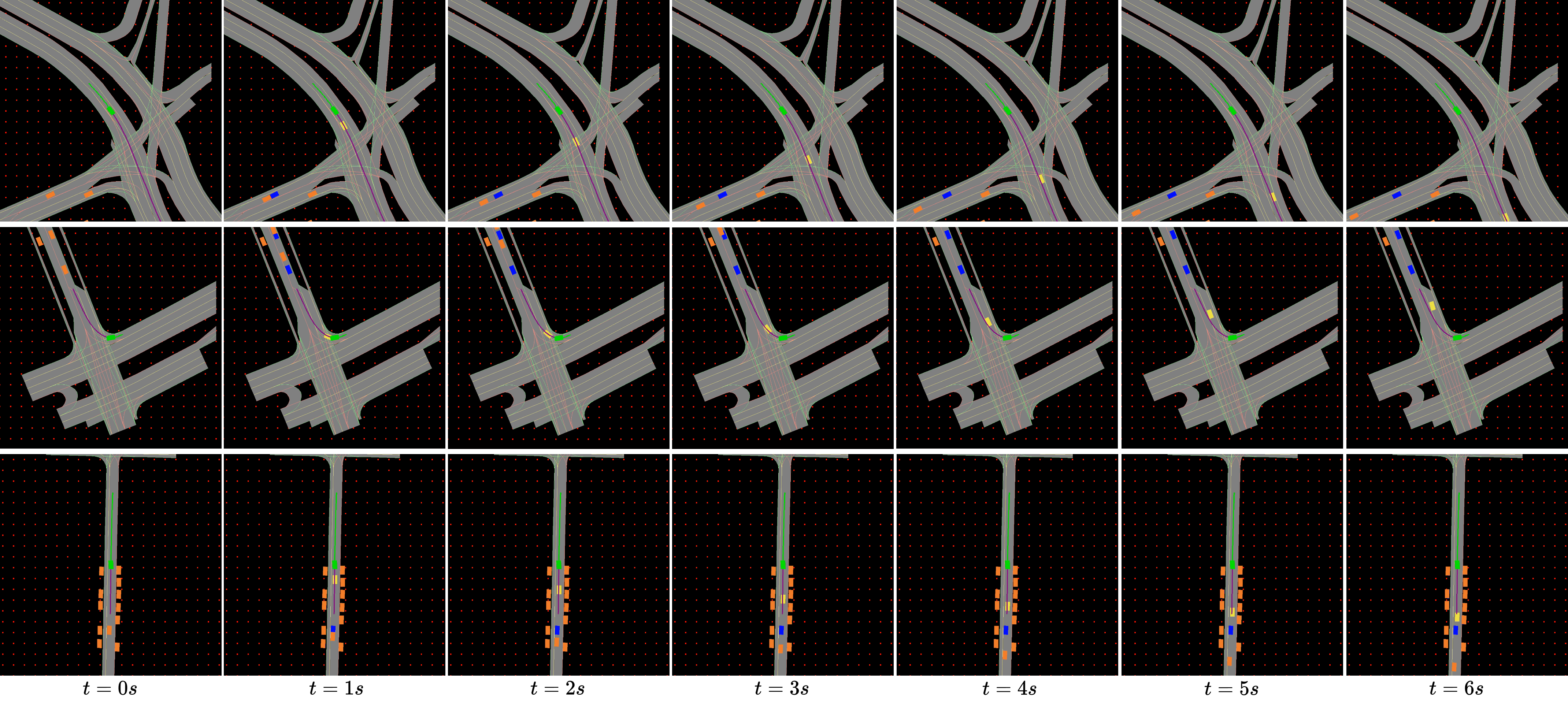}
    \vspace{-3mm}
    \caption{Six seconds of dreamed trajectories on the Argoverse 2 validation dataset. The green car represents the ego position at $t=0$ and the blue cars represent the dynamic objects at $t=0$. The purple line represents the ego's ground truth trajectory. The yellow and orange cars are the dreamed reconstructions of the ego and dynamic objects, respectively.}
    \label{fig:vrd_frames}
    \vspace{-5mm}
\end{figure*}

\vspace{-3mm}
\begin{table}[h!]
    \centering
    \caption{Ablation Study on Recurrent Model Parameters}
    \begin{tabular}{c|c|c|c|c}
        \hline\hline
        $T$ & $dt$ & $\mathrm{minADE_1}$ & $\mathrm{minFDE_1}$ & $\mathrm{actorMR_1}$ \\
         \hline
         2 & 0.1s & 1.68 & 4.17 & 0.37 \\
         2 & 0.2s & 1.86 & 4.61 & 0.38 \\
         2 & 0.5s & 1.87 & 4.61 & 0.37 \\
         4 & 0.1s & 1.88 & 4.66 & 0.38 \\
         4 & 0.2s & 1.89 & 4.65 & 0.37 \\
         4 & 0.5s & 2.01 & 4.98 & 0.40 \\
         \hline\hline
    \end{tabular}
    \begin{flushleft}
    Table \ref{tab:multidata} Quantitivave results on the Argoverse 2 dataset when the recurrent state-space model is trained using different horizon lengths and recurrent time step values.
    \end{flushleft}
    \label{tab:multidata}
    \vspace{-6mm}
\end{table}

 Table \ref{tab:multidata}. shows ablations where we study the effect of training the transition model for 30 epochs using different sequence lengths, $T$, and prediction time steps, $dt$, and evaluated on 7500 samples from the Argoverse 2 validation dataset. For evaluations on models with time step sizes greater than 0.1 seconds, we use the previously reconstructed kinematics to generate the trajectories at the intermediate time steps. The ablations show that there is not a significant difference in terms of inference performance between different sequence lengths. Models with a shorter prediction time step seem to perform better, but this is at the cost of longer training duration. This is because models with larger time steps train significantly quicker since there are fewer iterations when doing BPTT. However, the closed-loop dream has fewer opportunities to correct errors in the generated trajectories which results in larger average displacement errors. We also present quantitative results on both the inD and Argoverse 2 datasets for 3 and 6-second predictions in table \ref{tab:duration_comparison}.

\vspace{-3mm}
 \begin{table}[h!]
     \centering
     \caption{Evaluations on Different Data Distributions and Horizons}
     \begin{tabular}{l|c|c|c}
        \hline\hline
        \hspace{4mm} Dataset & $\mathrm{minADE_1}$ & $\mathrm{minFDE_1}$ & $\mathrm{actorMR_1}$ \\
        \hline
        InD (3s) & 0.14 & 0.37 & 0.07 \\
        Argoverse 2 (3s) & 0.68 & 1.41 & 0.22 \\
        InD (6s) & 0.49 & 1.39 & 0.12 \\
        Argoverse 2 (6s) & 1.68 & 4.17 & 0.37 \\
        \hline\hline
     \end{tabular}
     \vspace{1mm}
     \label{tab:duration_comparison}
     \begin{flushleft}
    Table \ref{tab:duration_comparison} Comparison of VRD on InD and Argoverse 2 datasets over different prediction horizons.
    \end{flushleft}
    \vspace{-5mm}
 \end{table}

The inD dataset contains more intersection scenarios than the Argoverse 2 dataset, allowing us to evaluate the performance of our model when the dynamic objects in the scenario are required to make turns. Furthermore, since this data is captured using a drone with a bird's eye view of each scenario, the data quality is better. Our model performs considerably well on the inD dataset demonstrating its ability to forecast dynamic objects on a variety of scenarios.
Fig. \ref{fig:vrd_frames} shows the first 3 seconds of dreamed trajectories of VRD on different scenarios of the Argoverse 2 dataset.

\vspace{-2mm}
\section{Conclusions \label{s5}}
\vspace{-2mm}
This paper introduces VRD, a world model based motion forecasting framework that proposes a new way to approach multi-agent motion forecasting. By leveraging a recurrent state-space model to learn transitions between observations, VRD dreams a realistic future of the environment over long horizons. By learning this powerful transition function we can propose a new training pipeline that combines traditional open-loop imitation learning with a dreaming-assisted closed-loop training regime. We demonstrate the effectiveness of VRD on the Argoverse 2 motion forecasting dataset, showing that our model achieves state-of-the-art performance on the single prediction miss rate metric and achieves comparable performance to the current top-performing models on the single prediction displacement error benchmarks.
\vspace{-4mm}

\bibliographystyle{IEEEtran.bst}

\typeout{}
\bibliography{reference.bib}

\begin{thebibliography}{10}
\providecommand{\url}[1]{#1}
\csname url@rmstyle\endcsname
\providecommand{\newblock}{\relax}
\providecommand{\bibinfo}[2]{#2}
\providecommand\BIBentrySTDinterwordspacing{\spaceskip=0pt\relax}
\providecommand\BIBentryALTinterwordstretchfactor{4}
\providecommand\BIBentryALTinterwordspacing{\spaceskip=\fontdimen2\font plus
\BIBentryALTinterwordstretchfactor\fontdimen3\font minus \fontdimen4\font\relax}
\providecommand\BIBforeignlanguage[2]{{%
\expandafter\ifx\csname l@#1\endcsname\relax
\typeout{** WARNING: IEEEtran.bst: No hyphenation pattern has been}%
\typeout{** loaded for the language `#1'. Using the pattern for}%
\typeout{** the default language instead.}%
\else
\language=\csname l@#1\endcsname
\fi
#2}}

\bibitem{Lee_desire}
N.~Lee, W.~Choi, P.~Vernaza, C.~B. Choy, P.~H.~S. Torr, and M.~K. Chandraker, ``{{DESIRE:} Distant Future Prediction in Dynamic Scenes with Interacting Agents},'' \emph{CoRR}, vol. abs/1704.04394, 2017.

\bibitem{Fang_tpnet}
L.~Fang, Q.~Jiang, J.~Shi, and B.~Zhou, ``{TPNet: Trajectory Proposal Network for Motion Prediction},'' \emph{CoRR}, vol. abs/2004.12255, 2020.

\bibitem{Lu_not_enough}
Y.~Lu, J.~Fu, G.~Tucker, X.~Pan, E.~Bronstein, R.~Roelofs, B.~Sapp, B.~White, A.~Faust, S.~Whiteson, D.~Anguelov, and S.~Levine, ``{Imitation Is Not Enough: Robustifying Imitation with Reinforcement Learning for Challenging Driving Scenarios},'' in \emph{2023 IEEE/RSJ International Conference on Intelligent Robots and Systems (IROS)}, 2023, pp. 7553--7560.

\bibitem{Knox_misdesign}
W.~B. Knox, A.~Allievi, H.~Banzhaf, F.~Schmitt, and P.~Stone, ``{Reward (Mis)design for Autonomous Driving},'' \emph{CoRR}, vol. abs/2104.13906, 2021.

\bibitem{Baram_MGAIL}
N.~Baram, O.~Anschel, I.~Caspi, and S.~Mannor, ``{End-to-End Differentiable Adversarial Imitation Learning},'' in \emph{Proceedings of the 34th International Conference on Machine Learning}, vol.~70.\hskip 1em plus 0.5em minus 0.4em\relax PMLR, 06--11 Aug 2017, pp. 390--399.

\bibitem{Bronstein_HMGAIL}
E.~Bronstein, M.~Palatucci, D.~Notz, B.~White, A.~Kuefler, Y.~Lu, S.~Paul, P.~Nikdel, P.~Mougin, H.~Chen, J.~Fu, A.~Abrams, P.~Shah, E.~Racah, B.~Frenkel, S.~Whiteson, and D.~Anguelov, ``{Hierarchical Model-Based Imitation Learning for Planning in Autonomous Driving},'' in \emph{2022 IEEE/RSJ International Conference on Intelligent Robots and Systems (IROS)}, 2022, pp. 8652--8659.

\bibitem{Caesar_Nuscenes}
H.~Caesar, V.~Bankiti, A.~H. Lang, S.~Vora, V.~E. Liong, Q.~Xu, A.~Krishnan, Y.~Pan, G.~Baldan, and O.~Beijbom, ``{nuScenes: A Multimodal Dataset for Autonomous Driving},'' in \emph{2020 IEEE/CVF Conference on Computer Vision and Pattern Recognition (CVPR)}, 2020, pp. 11\,618--11\,628.

\bibitem{WOMD}
S.~Ettinger, S.~Cheng, B.~Caine, C.~Liu, H.~Zhao, S.~Pradhan, Y.~Chai, B.~Sapp, C.~Qi, Y.~Zhou, Z.~Yang, A.~Chouard, P.~Sun, J.~Ngiam, V.~Vasudevan, A.~McCauley, J.~Shlens, and D.~Anguelov, ``{Large Scale Interactive Motion Forecasting for Autonomous Driving : The Waymo Open Motion Dataset},'' in \emph{2021 IEEE/CVF International Conference on Computer Vision (ICCV)}, 2021, pp. 9690--9699.

\bibitem{Zhan_Interaction}
W.~Zhan, L.~Sun, D.~Wang, H.~Shi, A.~Clausse, M.~Naumann, J.~Kummerle, H.~Konigshof, C.~Stiller, A.~de~La~Fortelle, and M.~Tomizuka, ``{INTERACTION Dataset: An INTERnational, Adversarial and Cooperative moTION Dataset in Interactive Driving Scenarios with Semantic Maps},'' 2019.

\bibitem{Argoverse2}
B.~Wilson, W.~Qi, T.~Agarwal, J.~Lambert, J.~Singh, S.~Khandelwal, B.~Pan, R.~Kumar, A.~Hartnett, J.~K. Pontes, D.~Ramanan, P.~Carr, and J.~Hays, ``{Argoverse 2: Next Generation Datasets for Self-driving Perception and Forecasting},'' in \emph{Proceedings of the Neural Information Processing Systems Track on Datasets and Benchmarks (NeurIPS Datasets and Benchmarks 2021)}, 2021.

\bibitem{Zhou_HIVT}
Z.~Zhou, L.~Ye, J.~Wang, K.~Wu, and K.~Lu, ``{HiVT: Hierarchical Vector Transformer for Multi-Agent Motion Prediction},'' in \emph{2022 IEEE/CVF Conference on Computer Vision and Pattern Recognition (CVPR)}, 2022, pp. 8813--8823.

\bibitem{Gao_HeteroGCN}
X.~Gao, X.~Jia, Y.~Li, and H.~Xiong, ``{Dynamic Scenario Representation Learning for Motion Forecasting With Heterogeneous Graph Convolutional Recurrent Networks},'' \emph{IEEE Robotics and Automation Letters}, vol.~8, no.~5, pp. 2946--2953, 2023.

\bibitem{Cheng_ForecastMAE}
J.~Cheng, X.~Mei, and M.~Liu, ``{Forecast-MAE: Self-supervised Pre-training for Motion Forecasting with Masked Autoencoders},'' in \emph{Proceedings of the IEEE/CVF International Conference on Computer Vision (ICCV)}, October 2023, pp. 8679--8689.

\bibitem{Zhou_QCNet}
Z.~Zhou, J.~Wang, Y.-H. Li, and Y.-K. Huang, ``{Query-Centric Trajectory Prediction},'' in \emph{Proceedings of the IEEE/CVF Conference on Computer Vision and Pattern Recognition (CVPR)}, 2023.

\bibitem{Ross_Dagger}
S.~Ross, G.~Gordon, and D.~Bagnell, ``{A Reduction of Imitation Learning and Structured Prediction to No-Regret Online Learning},'' in \emph{Proceedings of the Fourteenth International Conference on Artificial Intelligence and Statistics}, vol.~15, 11--13 Apr 2011, pp. 627--635.

\bibitem{Pomerleau_ALVINN}
D.~A. Pomerleau, ``{ALVINN: An Autonomous Land Vehicle in a Neural Network},'' in \emph{Advances in Neural Information Processing Systems}, vol.~1, 1988.

\bibitem{Salzmann_trajectron}
T.~Salzmann, B.~Ivanovic, P.~Chakravarty, and M.~Pavone, ``Trajectron++: Multi-agent generative trajectory forecasting with heterogeneous data for control,'' \emph{CoRR}, vol. abs/2001.03093, 2020.

\bibitem{Ngiam_scene}
J.~Ngiam, B.~Caine, V.~Vasudevan, Z.~Zhang, H.~L. Chiang, J.~Ling, R.~Roelofs, A.~Bewley, C.~Liu, A.~Venugopal, D.~Weiss, B.~Sapp, Z.~Chen, and J.~Shlens, ``{Scene Transformer: {A} unified multi-task model for behavior prediction and planning},'' \emph{CoRR}, vol. abs/2106.08417, 2021.

\bibitem{Keufler_GAN}
A.~Kuefler, J.~Morton, T.~A. Wheeler, and M.~J. Kochenderfer, ``{Imitating Driver Behavior with Generative Adversarial Networks},'' \emph{CoRR}, vol. abs/1701.06699, 2017.

\bibitem{Hu_Mile}
A.~Hu, G.~Corrado, N.~Griffiths, Z.~Murez, C.~Gurau, H.~Yeo, A.~Kendall, R.~Cipolla, and J.~Shotton, ``{Model-Based Imitation Learning for Urban Driving},'' in \emph{Advances in Neural Information Processing Systems (NeurIPS)}, vol.~35.\hskip 1em plus 0.5em minus 0.4em\relax Curran Associates, Inc., 2022, pp. 20\,703--20\,716.

\bibitem{Zhao_TNT}
H.~Zhao, J.~Gao, T.~Lan, C.~Sun, B.~Sapp, B.~Varadarajan, Y.~Shen, Y.~Shen, Y.~Chai, C.~Schmid, C.~Li, and D.~Anguelov, ``{TNT: Target-driven Trajectory Prediction},'' in \emph{Proceedings of the 2020 Conference on Robot Learning}, ser. Proceedings of Machine Learning Research, vol. 155.\hskip 1em plus 0.5em minus 0.4em\relax PMLR, 16--18 Nov 2021, pp. 895--904.

\bibitem{Yang_HDnet}
B.~Yang, M.~Liang, and R.~Urtasun, ``{HDNET: Exploiting HD Maps for 3D Object Detection},'' in \emph{Proceedings of The 2nd Conference on Robot Learning}, vol.~87, 29--31 Oct 2018, pp. 146--155.

\bibitem{Bansal_ChauffeurNet}
M.~Bansal, A.~Krizhevsky, and A.~Ogale, ``{ChauffeurNet: Learning to Drive by Imitating the Best and Synthesizing the Worst},'' in \emph{Proceedings of Robotics: Science and Systems (RSS)}, Germany, June 2019.

\bibitem{Liang_LaneGCN}
M.~Liang, B.~Yang, R.~Hu, Y.~Chen, R.~Liao, S.~Feng, and R.~Urtasun, ``{Learning lane graph representations for motion forecasting},'' in \emph{ECCV}, 2020.

\bibitem{Gao_Vectornet}
J.~Gao, C.~Sun, H.~Zhao, Y.~Shen, D.~Anguelov, C.~Li, and C.~Schmid, ``{VectorNet: Encoding HD Maps and Agent Dynamics From Vectorized Representation},'' in \emph{Proceedings of the IEEE/CVF Conference on Computer Vision and Pattern Recognition (CVPR)}, June 2020.

\bibitem{Ha_WorldModels}
D.~Ha and J.~Schmidhuber, ``{Recurrent World Models Facilitate Policy Evolution}, volume = {31},'' in \emph{Advances in Neural Information Processing Systems (NeurIPS)}, S.~Bengio, H.~Wallach, H.~Larochelle, K.~Grauman, N.~Cesa-Bianchi, and R.~Garnett, Eds., 2018.

\bibitem{Hafner_Learn2Control}
D.~Hafner, T.~P. Lillicrap, J.~Ba, and M.~Norouzi, ``{Dream to Control: Learning Behaviors by Latent Imagination},'' \emph{CoRR}, vol. abs/1912.01603, 2019.

\bibitem{Hafner_dreamerv2}
D.~Hafner, T.~P. Lillicrap, M.~Norouzi, and J.~Ba, ``{Mastering Atari with Discrete World Models},'' \emph{CoRR}, vol. abs/2010.02193, 2020.

\bibitem{Schrittwieser_muzero}
J.~Schrittwieser, I.~Antonoglou, T.~Hubert, K.~Simonyan, L.~Sifre, S.~Schmitt, A.~Guez, E.~Lockhart, D.~Hassabis, T.~Graepel, T.~Lillicrap, and D.~Silver, ``{Mastering Atari, Go, chess and shogi by planning with a learned model},'' \emph{Nature}, vol. 588, no. 7839, p. 604–609, Dec. 2020.

\bibitem{Zhang_Trafficbots}
Z.~Zhang, A.~Liniger, D.~Dai, F.~Yu, and L.~Van~Gool, ``{TrafficBots: Towards World Models for Autonomous Driving Simulation and Motion Prediction},'' in \emph{2023 IEEE International Conference on Robotics and Automation (ICRA)}, 2023, pp. 1522--1529.

\bibitem{inDdataset}
J.~Bock, R.~Krajewski, T.~Moers, S.~Runde, L.~Vater, and L.~Eckstein, ``{The inD Dataset: A Drone Dataset of Naturalistic Road User Trajectories at German Intersections},'' in \emph{2020 IEEE Intelligent Vehicles Symposium (IV)}, 2020, pp. 1929--1934.

\bibitem{Su_QML}
T.~Su, X.~Wang, and X.~Yang, ``{QML for Argoverse 2 Motion Forecasting Challenge},'' 2022.

\bibitem{think2drive}
Q.~Li, X.~Jia, S.~Wang, and J.~Yan, ``Think2drive: Efficient reinforcement learning by thinking in latent world model for quasi-realistic autonomous driving (in carla-v2),'' \emph{arXiv preprint arXiv:2402.16720}, 2024.

\end{thebibliography}

\end{document}